\documentclass[fleqn,twoside]{article}
\usepackage[headings]{espcrc2}

\readRCS
$Id: espcrc2.tex,v 1.2 2004/02/24 11:22:11 spepping Exp $
\usepackage{graphicx}
\usepackage[figuresright]{rotating}

\usepackage[dvips]{epsfig}
\usepackage{fancyhdr}

\newcommand{\AmS}{{\protect\the\textfont2
  A\kern-.1667em\lower.5ex\hbox{M}\kern-.125emS}}

\newcommand{\cm}{\mathrm{cm}}

\title{Evaluation of Uncertain Image Classification and Segmentation Algorithms }

 \author{Arnaud Martin\address[E3I2]{ENSIETA E3I2 EA3876, \\
        2 rue Fran{\c c}ois Verny, 29806 Brest Cedex 09, France},
        Hicham Laanaya\addressmark  \address[MV]{Facult{\'e} des sciences de Rabat, \\
        Avenue Ibn Batouta, B.P. 1014 Rabat, Morocco}
       and
        Andreas Arnold-Bos\addressmark[E3I2]} 

\runtitle{Evaluation of Uncertain Image Classification and Segmentation Algorithms}
\runauthor{A. Martin et al.}

\begin{document}

\begin{abstract}
Each year, numerous segmentation and classification algorithms are invented or reused to solve problems where machine vision is needed. Generally, the efficiency of these algorithms is compared against the results given by one or many human experts. However, in many situations, the location of the real boundaries of the object as well as their classes are not known with certainty by the human experts. Moreover, only one aspect of the segmentation and classification problem is generally evaluated. In our evaluation method, we take into account both the classification and segmentation results as well as the level of certainty given by the experts. As a concrete example of our method, we evaluate an automatic seabed characterization algorithm based on sonar images. 
\vspace{1pc}
\end{abstract}

\maketitle

\section{INTRODUCTION}

Image classification and segmentation are two fundamental problems in image analysis. Segmenting an image consists in dividing the image into homogeneous zones delimited by boundaries so as to separate the different entities visible in the image. Classification consists in labeling the various components visible in an image. A great deal of segmentation and classification methods have been proposed in the last thirty years \cite{Russ02}; enumerating them all is not the purpose of our paper. However, an important question to solve is how to benchmark these methods and evaluate their robustness with respect to a given real-life application.

A typical example of the use of classification and segmentation is encountered in satellite or sonar imaging, where an important use of the data is to classify the types of soils present in the images, for instance to build maps. As the amount of images gathered during a mission is important, automatic recognition algorithms can relieve human operators. Since the swath of the sensor is wide, many types of soils can be encountered within a single image, and the classification must be done on a local neighborhood. This neighborhood can be either limited to a single pixel, or often to a small tile of {\it e.g.} 16 $\times$ 16 or 32 $\times$ 32 pixels taken as the unit for the classification algorithm. The boundaries between the different patches corresponding to a category of soil are a form of segmentation, which is here an implicit byproduct of the classification. In other applications, segmentation can come first so as to isolate entities which will be labeled later.

\begin{figure}[htb]
\vspace{-0.5cm}
\includegraphics[height=5cm]{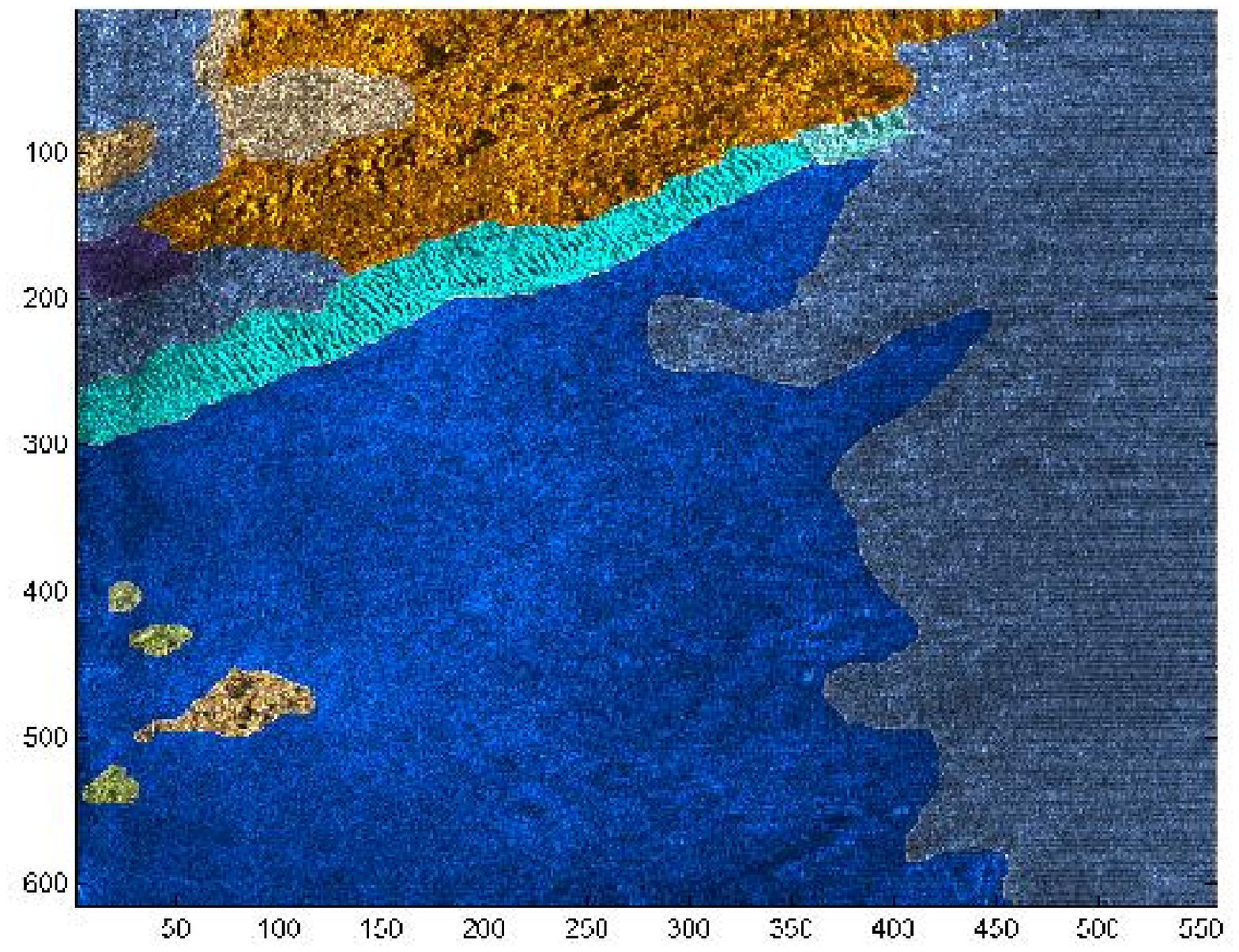}
\includegraphics[height=5cm]{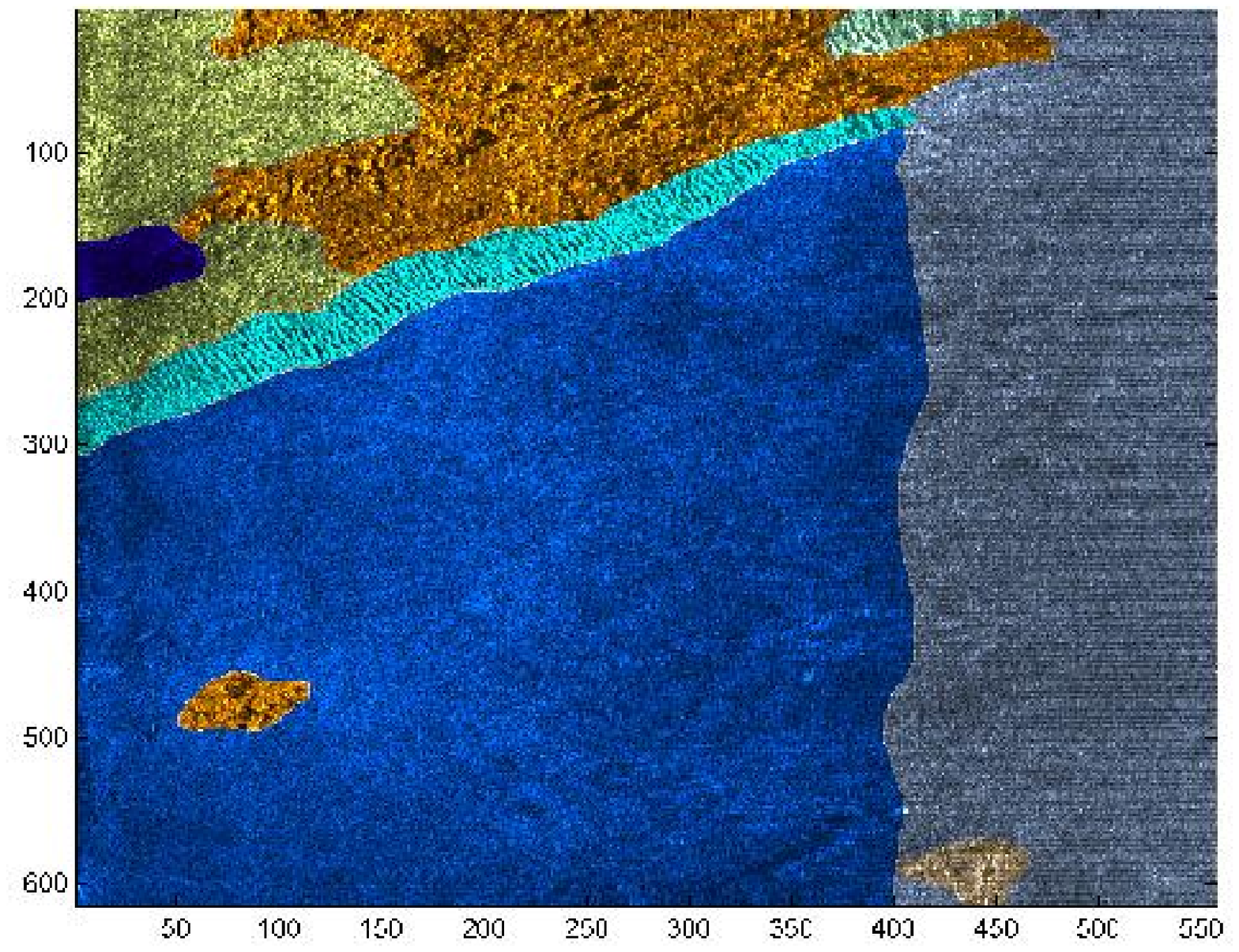}
\vspace{-0.5cm}
\caption{Segmentation given by two experts. \label{expert}}
\vspace{-0.5cm}
\end{figure}

A difficulty raised in these applications is the lack of ground truth which could be used to evaluate the result of the classification. The real reference classes must be estimated by human experts from the data themselves. However, the images are difficult to read since they are corrupted by many phenomena and the estimation of the classes by the human expert will be highly subjective and with a varying level of uncertainty. In the case of the automatic seabed classification, which we will use as our reference example throughout this paper, images are especially hard to interpret due to many imperfections \cite{Martin05}. To reconstruct the image, a huge number of parameters (geometry of the device, coordinates of the ship, movements of the sonar, etc.) are taken into account, but these data are polluted with a large amount of sensor noise. Plus, other phenomena such as multipath signal propagation (caused by reflection either on the bottom or the surface), speckle, and the presence of fauna and flora (\emph{e.g.} shadows of fishes on the sea bottom), will all augment the difficulty of interpretation of the image. Consequently, different experts can propose different classifications of the image. Thus, in order to evaluate automatic classification, we must take into account this difference and the uncertainty of each expert. Figure \ref{expert} exhibits the differences between the interpretation and the certainty of two sonar experts trying to differentiate the type of sediment (rock, cobbles, sand, ripple, silt) or shadow when the information is invisible (each color correspond to a kind of sediment and the associated certainty of the expert for this sediment expressed in term of sure, moderately sure and not sure). 

We propose in this article a new approach for image classification and segmentation taking into account the information giving by multiple experts and the certainty of the given information. Classical evaluations of the classification and segmentation do not take into account the uncertain and imprecise labels in the reference image provided by an expert. We think that we have to consider these kind of labels in our evaluation approach. In section \ref{classification} we show how to integrate the expert certainty in confusion matrix and so to deduce a good classification rate and error classification rate. Moreover, our thesis is that global image classification evaluation must be made not only by evaluating the classification on considered units (with the confusion matrix) but also by evaluating, at the same time, the induced segmentation. In section \ref{segmentation}, we propose two new distance-based measures in order to evaluate well and mis-segmented pixels by taking into account both the location of the borders and the expert certainty. Note that another important criterion to evaluate classification/segmentation approaches is the evaluation of the complexity of the algorithms \cite{Zhang96}, but we do not consider it in this paper.
Finally, our evaluation is illustrated in section \ref{illustration} on real sonar images acquired in a real, uncertain environment.

\section{CLASSIFICATION EVALUATION}
\label{classification}

Traditional classification systems can usually be described as a three-tiered process. First, significant features are extracted from the images to classify. These features are widely different, depending on the application; they are generally described using a small set of abstract numerical measures. For example, used features may be the local luminance, the texture (described with measures such as the entropy, the co-ocurrence matrices, etc), the contours (described with their length, their orientation, their relative position to other contours, etc) \cite{Russ02}. Most of the time, a second stage is necessary to reduce these features, because there are too numerous. In the third stage of the algorithm, the numerical descriptors are fed to classification algorithms, which are application-independent, such as Support Vector Machine \cite{Laanaya05,David05,Mavroforakis05}, neural networks \cite{Martin05,Mavroforakis05,Park04,Venkatesh03}, $k$-nearest neighbors \cite{Leblond05}, etc. The classification algorithms will decide, depending on their entries, which is the class of the image. 

Hence, we have to evaluate these classification algorithms in order to compare their robustness in a given application. The classical approach is based on the confusion matrix and does not take into account uncertain labels. We propose here a new confusion matrix and good classification and error rates taking into account these kind of labels and also the inhomogeneous units defined forwards.

The proposed method of evaluation in this section, can be applied for the evaluation of a classification algorithm in every domain where uncertain labels are provided. We do not consider here the problem of the learning on uncertain and imprecise labels \cite{Pizzi99,Denoeux00,Vannoorenberghe05}: the classification can be made by this kind of algorithms or others.

\subsection{Classical Evaluation}

The results of one image classification can be observed and visually compared to the reality. But in order to evaluate a classification algorithm, many different configurations and tests must be considered. Classification algorithms can yield very variable results depending on the sample. Generally classification algorithms evaluation is conducted by the confusion matrix. Confusion matrix is composed by the number $\cm_{ij}$ of elements from the class $i$ classified in the class $j$. In order to obtained rates, with one more easier to interpret, we can normalize this confusion matrix by:
\begin{equation}
\label{NCM}
Ncm_{ij} = \frac{\cm_{ij}}{\sum_{j=1}^N \cm_{ij}}=\frac{\cm_{ij}}{N_i}, 
\end{equation}
with $N$ the number of considered classes and $N_i$ the number of element from the true class $i$. From this normalized confusion matrix a good classification rate vector can be written as:
\begin{equation}
\label{GCR}
GCR_i = Ncm_{ii}, 
\end{equation}
and an error classification rate vector as:
\begin{equation}
\label{ECR}
ECR_i = \frac{1}{2} \left(\sum_{j=1, j\neq i}^{N} Ncm_{ij}+  \sum_{i=1, i \neq j}^{N} \frac{Ncm_{ij}}{N-1} \right). 
\end{equation}

This error classification rate is the mean of the two errors corresponding respectively to the elements from a given class $i$ falsely classified as elements of another class (first term), and to the elements classified in a given class $j$ but being from another class $i$ (second term). These errors are also called errors of first and second kind. We do not have to normalize the first term because of the normalization of the confusion matrix on the rows, but the second term must be normalized by the number of rows minus one (because of the $Ncm_{ii}$ term corresponding to the good classification). Note that other error rates can be defined (see {\em e.g.} \cite{Pizzi99}).

We have seen that image classification algorithms evaluation must be made not only on one image but on the whole image database. As a trivial consequence, we have to consider a non-normalized confusion matrix on each image and normalize the sum of the matrix confusion on all images of the database. 

\subsection{Evaluation with expert information}

Consider a general case where information is given by the expert on each pixel and the classification algorithm is made on an unit of $n \times n$ pixels. In such a case, if a $n \times n$ tile is considered, more than one class can be present (we call it patch-worked tile or inhomogeneous unit), and the classification algorithm can find only one of these class. In order to take into account the last example, we consider that if the classification algorithm finds one of these classes on the tile, the algorithm is right in the proportion of this class found in the $n \times n$ tile and it is false in the proportion of the other classes in the tile. For instance, imagine the case where the expert considers a $16 \times 16$ tile and declares that 156 given pixels belong to class 1, and 100 other pixels belong to class 3. If the classification algorithm finds the tile belongs to class 1, the confusion matrix will be computed by $\cm_{11}=\cm_{11}+156/256$ and $\cm_{31}=\cm_{31}+100/256$. Hence the confusion matrix is not composed of integer numbers and $N_i$ is also not integer, but the sums of column are still integer.

Now consider the case where the expert gives the class with a certainty grade. For instance, the operator can be moderately sure of his choice when he labels one part of the image as belonging to a certain class, and be totally doubtful on another part of the image. In our classification evaluation we must not take these two references equally. Indeed, classical confusion matrices imply that the reality is perfectly known; this, unfortunately, is not the case in many real applications. We propose to represent this difference of information by different weights corresponding to the different certainty grades that are considered. For example, if three grades of certainty (sure, moderately sure and not sure) are considered, we can provide respectively the weights: 2/3, 1/2 and 1/3. In the confusion matrix, such weights could be integrated easily in the general sum. If one expert labels a tile as belonging to class 1, with a moderate certainty, and if the classification algorithm finds the class 1, considering the previous given weights, the confusion matrix will be updated such as:  $\cm_{11}=\cm_{11}+1/2$. If the classification algorithm finds the class 2 on the considered tile, the confusion matrix becomes $\cm_{12}=\cm_{12}+1/2$. Hence the sums of column are not integer anymore.

In order to take into account the referenced images provided by different experts, we can compare the classified image with all the expert-referenced images. Hence we obtain as many confusion matrices as experts, and we can simply combine them by addition. 

By the simple fact that we add the non-normalized confusion matrices, we weight the obtained results by the image size or the considered unit number.

Consequently, in order to obtain rates, we can normalize the obtained confusion matrix with equation (\ref{NCM}) and calculate the good classification rate vector with equation (\ref{GCR}) and the error classification rate vector with equation (\ref{ECR}). Of course these rates are not percentages anymore. For instance, the good classification rate is not percentage of well classified units anymore, because the weights given by the inhomogeneous units or by expert certainty are rational. 

In conclusion of this section: the interest of these newly obtained confusion matrix, good classification rate and error classification rate is that, they give a good evaluation of classification taking into account the inhomogeneous units and uncertainty of the experts. This approach can be applied in other applications than image classification, in fact in every domain where we try to classify uncertainty elements.

\section{SEGMENTATION EVALUATION}
\label{segmentation}

Segmentation can either be obtained as a byproduct of the classification, as shown above, or be used as the first step of an image processing pipeline. Many methods of image segmentation and edge detection have been proposed \cite{Canny86,Kovesi99,Alata05,Mena05,Bhalerao01}. It is important to be able to benchmark these methods and to evaluate their robustness; but to do that, measures are needed so as to have an objective means to judge the quality of the segmentation. No perfect measure exists today, and existing measures are not well satisfied, this is why we can imagine fusing the segmentation evaluation approaches \cite{Zhang05}.

On the one hand the image classification methods are evaluated by the confusion matrix. Good classification rates and error rates are usually calculated from this matrix. Note that in order to establish the confusion matrix, the real class of the considered units of the images need to be known. This gives only an evaluation of the classification approach on considered units of the image, but does not give an evaluation of the produced segmentation.

On the other hand, segmentation evaluation cannot be made only by visual comparison between the initial image and the segmented image. Many evaluation approaches have been proposed for image segmentation \cite{Zhang96,Mena05,Zhang97,Yitzhaky03,Roman01}. We can consider two cases: we do not have any {\it a priori} knowledge of the correct segmentation, and we have an {\it a priori} knowledge of the correct segmentation. In the first case, many effectiveness measures based on intra-region uniformity, inter-region contrast and region shape have been proposed \cite{Zhang96}. The second case implies to get referenced images. In a real application, experts must manually provide the image segmentation {\it via} a visual inspection. \cite{Zhang96} gives a review of usual discrepancy measures based on different distances (sometimes expressed in terms of probability) between the segmented-pixel and the referenced-pixel. 

Most of the time, only a measure of how many pixels are mis-segmented is given. We, on the contrary, propose in this article a combined study of one well-segmented pixel measure and a mis-segmented pixel measure. Indeed, most of the time, when a pixel is not mis-segmented, it is not necessary well-segmented either. As a consequence, we can have few mis-segmented pixels but also few well-segmented pixels, which means that the segmentation is not good overall.

In order to calculate confusion matrices we need the {\it a priori} knowledge of the class for each pixel or at least for each considered unit of the image. Hence, experts have to give referenced images, and we can consider to be in the second case of segmentation evaluation that we described above. 

Before presenting our method of segmentation evaluation, we show how we can easily obtain a deducted segmentation from an image classification based on the classification on tiles. Next, the proposed segmentation evaluation method is adapted to every image segmentation and can take into account imprecise labels.

\subsection{Deducted segmentation}
Image classification provides an implicit image segmentation given by the difference of classes between two adjacent tiles. Hence a good image classification evaluation should take this segmentation into account as well. 

First of all, we have to define the boundary pixels given by the image classification. We propose here to use a very simple approach: we will take as boundary pixels, the pixels which neighbor another class on the right or/and on the bottom. For instance, on table \ref{egSeg} we give a dummy segmented image with two classes given by $\times$ and $\bullet$. The classification unit is here 4 $\times$ 4. The boundary pixels are underlined. 

\begin{table}[htb]
\caption{Example of an obtained segmentation on image with two classes given by $\times$ and $\bullet$.}
\label{egSeg}
\begin{center}
\begin{tabular}{cccccccc}
$\times$  & $\times$  & $\times$  & \underline{$\times$} & $\bullet$ & $\bullet$ & $\bullet$ & $\bullet$ \\
$\times$  & $\times$  & $\times$  & \underline{$\times$} & $\bullet$ & $\bullet$ & $\bullet$ & $\bullet$ \\
$\times$  & $\times$  & $\times$ & \underline{$\times$} & $\bullet$ & $\bullet$ & $\bullet$ & $\bullet$ \\
\underline{$\times$}  & \underline{$\times$} & \underline{$\times$}  & \underline{$\times$} & \underline{$\bullet$} & \underline{$\bullet$}& \underline{$\bullet$} & \underline{$\bullet$} \\
$\bullet$ & $\bullet$ & $\bullet$ & \underline{$\bullet$} & $\times$ & $\times$ & $\times$ & $\times$ \\
$\bullet$ & $\bullet$ & $\bullet$ &\underline{$\bullet$} & $\times$  & $\times$ & $\times$ & $\times$ \\
$\bullet$ & $\bullet$ & $\bullet$ &\underline{$\bullet$} & $\times$  & $\times$ & $\times$ & $\times$ \\
$\bullet$ & $\bullet$ & $\bullet$ &\underline{$\bullet$} & $\times$  & $\times$ & $\times$ & $\times$ \\
\end{tabular}\\
\end{center}
\end{table}

Many approaches can be considered in order to obtain boundaries without angular points. We can consider for instance an interpolation between the 4-connexity or 8-connexity points \cite{Bouet00}. This is not the subject of this paper; the reader should keep in their mind that our segmentation evaluation is general and can be applied to all image segmentations given by boundary pixels.

\subsection{Segmentation evaluation}
We recall here that in our case, we have an {\it a priori} knowledge of the correct or approximately correct segmentation given by the experts. In this case all evaluation approaches are based on different distances (or probabilities) between the segmented-pixel and the referenced-pixel \cite{Zhang96,Peli82,Kanungo95} and most of the time only one measure of mis-segmented pixel is given. We think that it is not enough for a precise segmentation evaluation if a pixel can be not mis-segmented, and also not well-segmented. As we mentioned before, we can have few mis-segmented pixels only with few well-segmented pixels, and so the segmentation cannot be considered right. So we propose a linked study of two new measures: one well-segmented pixel measure and one mis-segmented pixel measure. Moreover these two measures can take into account the uncertainty of the expert on the position and on the existence of the boundaries if this uncertainty can be expressed as a weight.

\subsubsection{Boundary good detection measure}
\begin{figure}[htb]
\vspace{-0.5cm}
\centerline{
\includegraphics[width=7cm]{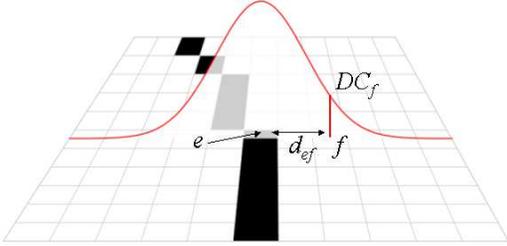}
} 
\vspace{-0.5cm}
\caption{Distance weight for the well-detection criteria.}
\vspace{-0.5cm}
\label{fig_pond}
\end{figure}

The well segmented pixel measure is a measure of how the boundary is well detected and the mis-segmented pixel measure tries to quantify how many boundaries detected by the algorithm to benchmark have no physical reality. First, we search the minimal distance $d_{fe}$ between each boundary pixel $f$ found by the algorithm to benchmark, and all the boundary pixels $e$ provided by the expert. Hence the pixel $e$ is a function of $f$, and we should note it as $e_f$, but in order to simplify notations, it is referred as $e$ in the rest of paper. We take here an Euclidean distance but any other distance can be envisaged. The certainty weight of the pixel $e$ given by the expert is noted as $W_e$. We define a well-detection criteria vector by:
\begin{equation}
\label{DC}
DC_f = \exp(-(d_{fe}*W_e)^2)*W_e. 
\end{equation}
This criteria gives a Gaussian-like distribution of weights with a standard deviation given by the certainty weights as shown in figure \ref{fig_pond}.

The boundary good detection measure is defined by the normalized well-detection criteria given by:
\begin{equation}
\label{WDC}
WDC = \frac{\sum_f DC_f} {\left(\max_f(DC_f) * \sum_e W_e\right)^a}.
\end{equation}
The normalization is made in order to obtain a measure defined between 0 and 1. However, in real applications, this criteria remains small even for very good boundary detection. So we take $a=1/6$ in order to accentuate small values. 

This criteria is not completely satisfying because it only takes into account the {\it distance} from the found boundary to the contour provided by the expert. However, the reference boundary also has a local {\it direction} which is another information we want to use. A boundary found by the algorithm can come across a boundary given by the expert orthogonally: in this case some pixels from the found boundary are very near (in terms of distance) to pixels from the given boundary but we do not want say that is a good detection. We propose two ways to consider the direction of boundaries. 

\begin{figure}[htb]
\vspace{-0.5cm}
\centerline{
\includegraphics[width=4cm]{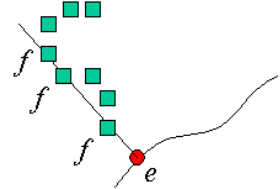}
} 
\vspace{-0.5cm}
\caption{Example of $n_{ef}$ for three given $f$, the found boundary is represented by green square and the referenced boundary by black line.}
\vspace{-0.5cm}
\label{fig_n_ef}
\end{figure}

In the first one, we count, for a given pixel $f$ of the found boundary, how many pixels from the found boundary are linked by the minimal distance to the same pixel $e$ of the referenced boundary. This number is noted $n_{ef}$, {\em e.g.} on figure \ref{fig_n_ef} we have $n_{ef}=3$ for three different $f$. We redefine the well-detection boundary measure by:
\begin{equation}
\label{WDCN}
WDC = \frac{\sum_f DC_f/n_{ef}} {\left(\max_f(DC_f/n_{ef}) * \sum_e W_e\right)^a}.
\end{equation}

The problem is that the number $n_{ef}$ does not necessarily represent a number of pixels on the same boundary and takes well into account only the orthogonal direction. However this measure gives the best evaluation of the proportion of the found boundaries.

The second method is based on the idea that the local direction of the boundary should also be taken into account: the direction of the detected boundary and the direction of the boundary given by the expert should be the same. Now, how does one compute the direction of the boundary? Let $I_r$ denote the reference boundary image given by the expert. $I_r(i,j)=0$ if no boundary is detected at pixel $(i,j)$; $I_r(i,j) = W_e$ otherwise, where $W_e$ is the weight of the pixel boundary $e$ at location $(i,j)$ given by the expert. Image $I_r$ can be seen as a discrete 2-D function on which the gradient $\overrightarrow g_r = [\partial I_r/\partial x;  \partial I_r/\partial y]$ can be computed. The gradient has the property to be normal to iso-values lines of $I_r$ and will therefore be normal to the boundaries given by the expert. Similarly, one can also compute the gradient $\overrightarrow g_s$ of the found boundary image. Then, a measure of correspondence between the directions at pixel $(i,j)$ can be given by the absolute value of the normalized dot product between the two gradients vectors\footnote{The notation ``.'' for multiplication is a term by term multiplication of the two matrices.}:

\begin{equation}
\label{eq_dotp}
BD = \frac{|\overrightarrow g_r.\overrightarrow g_s|}{|| \overrightarrow g_r||.||\overrightarrow g_s ||}.
\end{equation}

However, as $I_r$ is mostly filled with zeros, the gradient will have a negligible value at most locations. The farther a pixel is from a boundary given by the operator, the lower the gradient at that pixel will be, thus yielding a huge imprecision on the local direction of the image. To solve this problem, we used the Gradient Vector Flow (GVF), first introduced by Xu and Prince \cite{Xu98}. For a boundary image $I$, the GVF is a vector field $\overrightarrow f = [u(x,y);v(x,y)]$ that is computed iteratively so as to minimize the following cost function over all the boundary image:

\begin{eqnarray}
U = \int \!\!\! \int \left( \mu.(u_x^2+u_y^2+v_x^2+v_y^2)+ \ldots \right. \nonumber \\
    \left. + ||\overrightarrow g||^2|| \overrightarrow g-\overrightarrow f||^2 \right).dx.dy.
\end{eqnarray}
where $\mu$ is a tunable weight, variables in indices denote partial derivation with respect to that variable, and $\overrightarrow g$ is the gradient of the image as defined previously. This cost function was devised so that on boundaries, where the gradient is high ($||\overrightarrow g|| \rightarrow \infty$) the energy remains bounded: $||\overrightarrow g - \overrightarrow f ||$ must tend to zero if one wishes the integrand to be minimized. Thus, \emph{on boundaries, the GVF is equal to the gradient field}. On the other hand, for pixels far away from an  y boundary, the gradient will tend toward zero, and the integrand will be driven by the term $\mu.(u_x^2+u_y^2+v_x^2+v_y^2)$. To minimize it, the partial derivatives of the vector field $\overrightarrow f$ must be null, which means that \emph{the GVF extends the gradient by continuity to zones where it would normally be negligible}. The GVF is computed both for the reference image and the image obtained through segmentation. The measure of correspondence between the boundary directions will be similar to equation (\ref{eq_dotp}):

\begin{equation}
\label{eq_dotp2}
BD = \frac{|\overrightarrow f_r.\overrightarrow f_s|}{|| \overrightarrow g_r||.||\overrightarrow g_s ||}.
\end{equation}

\begin{figure}[htb]
\vspace{-0.5cm}
\begin{center}
\includegraphics[height=5cm]{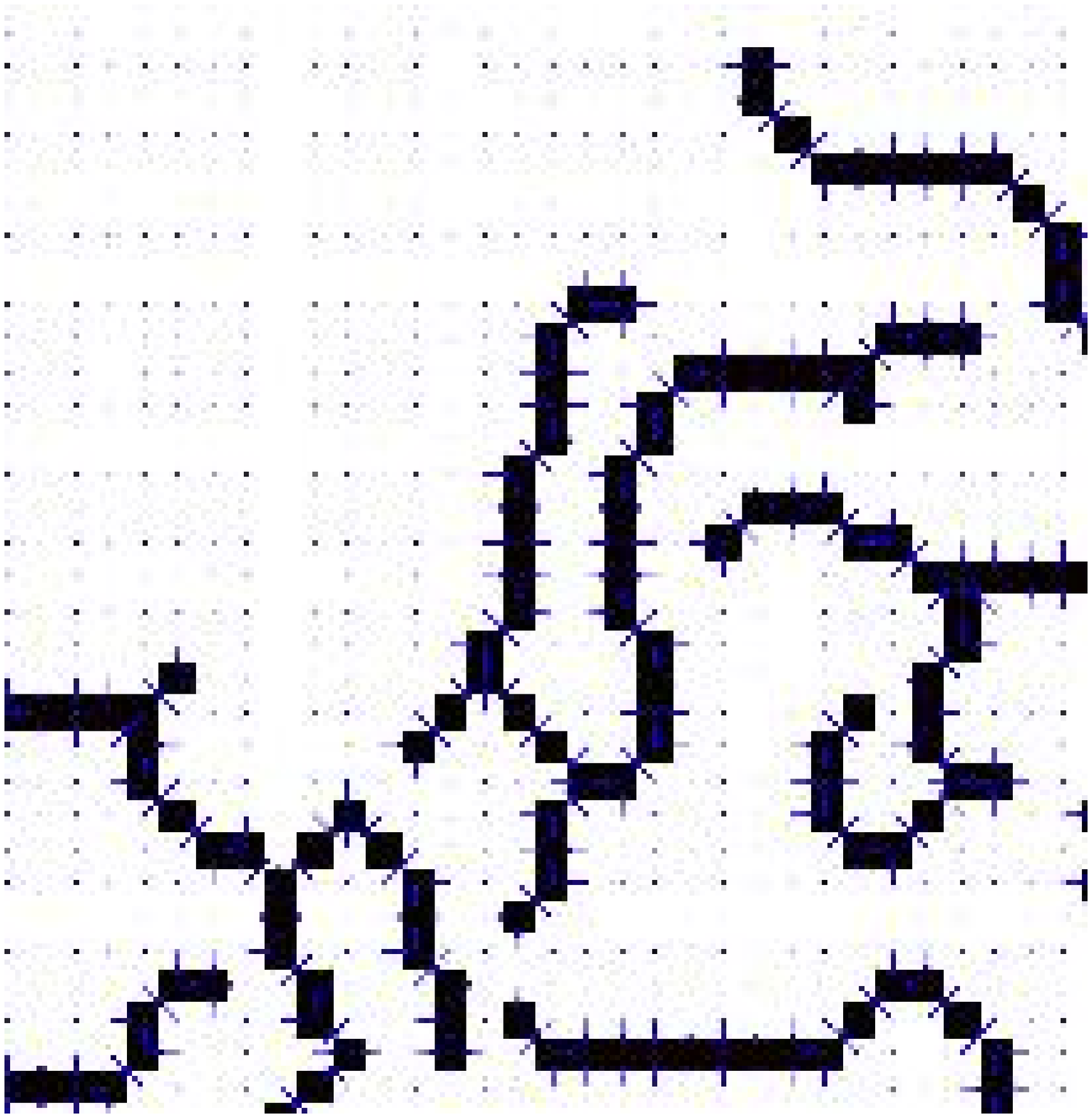}
\end{center}

\begin{center}
\includegraphics[height=5cm]{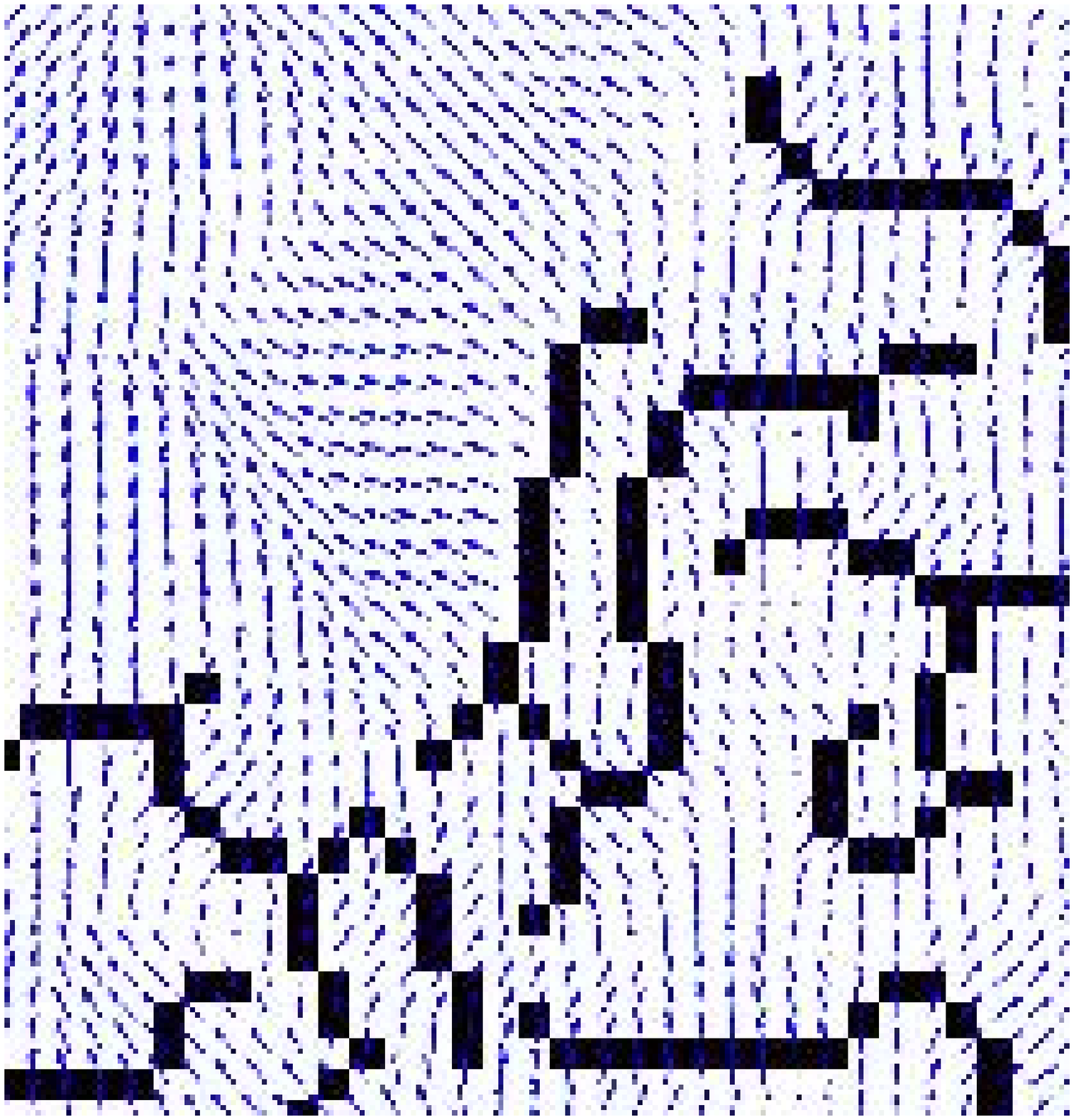}
\end{center}
\vspace{-0.5cm}
\caption{Computing the direction of the boundaries: gradient (top), GVF (bottom).}
\vspace{-0.5cm}
\label{direction}
\end{figure}

On figure \ref{direction}, note that the gradient is only strong on edges, whereas the GVF is strong everywhere, thus enabling the local directions to be seen.

Hence, we can redefine $DC_f$ in equation (\ref{WDCN}) by $(DC.BD)_f$, so that we obtain a new measure which takes into account the local direction of the found boundaries.

\subsubsection{False detection boundary measure}
The boundary false detection measure is based on the same principle than the well-detected boundary measure, but the Gaussian-like distribution of weights must be inversed. Hence we can defined a false detection criteria by:
\begin{equation}
\label{FDC}
FDC_f = 1-DC_f / W_e, 
\end{equation}
where the pixels $f$ and $e$ are linked by the minimal distance $d_{fe}$. As a consequence, the false detection boundary measure can be defined by the normalized false detection criteria by:
\begin{eqnarray}
\label{FD}
FD = 1- & \exp \left( -\frac{\sum_f\left(FDC_f*n_{ef}\right)}{\max_f(FDC_f*n_{ef}) * \sum_e W_e}\right).
\end{eqnarray}

In order to take into account the local direction of the found boundaries as found with the GVF, we can redefine $DC_f$ in equation (\ref{WDCN}) by $(FDC.(1-BD))_f$, so we obtain another new false detection criteria. 
 
Here we have described the use of measures $FD$ and $WDC$ for one image classified by the algorithm and another image provided by only one expert. In order to evaluate image segmentation algorithms on many images we can use a weighted sum of these both measures, taking into account the image sizes, which can be different for all considered images.

In conclusion of this section, we have described two new measures $FD$ and $WDC$ taking into account the uncertainty of different experts on the seen boundaries. We have to consider these two measures together.

\section{ILLUSTRATION}
\label{illustration}
We present here an illustration of our image classification and segmentation evaluation on real sonar images. Indeed, underwater environment is a very uncertain environment and it is particularly important to classify seabed for numerous applications such as Autonomous Underwater Vehicle navigation. In recent sonar works ({\it e.g.} \cite{LeChenadec05,Lianantonakis05}), the classification evaluation is made only by visual comparison of one original image and the classified image. That is not satisfying in order to correctly evaluate image classification and segmentation.
First we present our database given by two different experts with different certainties. Then, one possible classification approach for sonar image is presented. Finally the automatic classification and segmentation obtained by this approach is evaluated with our new evaluation method. 

Note that this illustration is presented in order to show how our measures work on only one classifier. In order to evaluate a classifier, we have to compare the results with another classifier or with other parametrization of the evaluated classifier.

\subsection{Database}
Our database contains 42 sonar images provided by the GESMA (Groupe d'Etudes Sous-Marines de l'Atlantique). Theses images were obtained with a Klein 5400 lateral sonar with a resolution of 20 to 30 cm in azimuth and 3 cm in range. The sea-bottom depth was between 15 m and 40 m.

The experts have manually segmented these images giving the kind of feature visible in a given part of the image: sediment (rock, cobble, sand, silt, ripple -either horizontal, vertical or at 45 degrees), shadow or other features (typically shipwrecks). All sediments are given with three certainty levels (sure, moderately sure or not sure), and the boundary between two sediments is also given with a certainty (sure, moderately sure or not sure). Hence, every pixel of every image is labeled as being either a certain type of sediment or a shadow, or a boundary with one of the three certainty levels. Figure \ref{expert} gives an example of such a segmentation provided by the expert.

\subsection{Classification approach}
The classification approach is based on supervised classification. In order to teach the classifier we have randomly divided the database into two parts. On the learning database we have considered, on randomly chosen images only, the homogeneous tiles with a 32 $\times$ 32 size and with a sure or moderately sure certitude level until to get approximately the same number of tiles in the learning and test databases. On the test database we have considered tiles with a 32 $\times$ 32 size and a recovering step of 4. On each tile we have extracted some features by a wavelet decomposition.

The discrete translation invariant wavelet transform is based on the choice of the optimal translation for each decomposition level. Each decomposition level $d$ gives four new images. We choose here a decomposition level $d=2$. For each image $I_d^i$ (the $i^{th}$ image of the decomposition $d$) we calculate three features. The energy is given by:
\begin{equation}
\label{energy}
\frac{1}{NM}\sum_{n=1}^N \sum_{m=1}^{M}I_d^i(n,m),
\end{equation}
where $N$ and $M$ are respectively the number of pixels on the rows, and on the columns. The entropy is estimated by:
\begin{equation}
\label{entropy}
-\frac{1}{NM}\sum_{n=1}^N \sum_{m=1}^{M}I_d^i(n,m) \ln(I_d^i(n,m)),
\end{equation}
and the mean is given by:
\begin{equation}
\label{mean}
\frac{1}{NM}\sum_{n=1}^N \sum_{m=1}^{M}|I_d^i(n,m)|.
\end{equation}
Consequently we obtain 15 features (3+4*3).

The chosen classifier is based on a Support Vector Machine. The algorithm used here is described in \cite{Chang01}. It is a one-{\it vs}-one multi-class approach, and we take a linear kernel with a constant $C=1$.

We have considered only three classes for learning and tests: 
\begin{itemize}
\item [-] class 1: Rock and Cobble
\item [-] class 2: Ripple in all directions
\item [-] class 3: Sand and Silt
\end{itemize}
Hence shadow is not considered and so the classification can not be good on tiles with shadow. In order to take into account unknown classes, one solution is to add a rejected class in the classifier. However, as we show farther down, we can also take into account this class if the classifier has no rejected class. 

The units of the classifier are tiles with a 32 $\times$ 32 size with a recovering step of 4. Hence, we can classify tiles with a 4 $\times$ 4 size, considering the tile of 4 $\times$ 4 size in the middle on each tile of 32 $\times$ 32.

\subsection{Evaluation}

\begin{figure}[htb]
\vspace{-0.5cm}
\includegraphics[height=5cm]{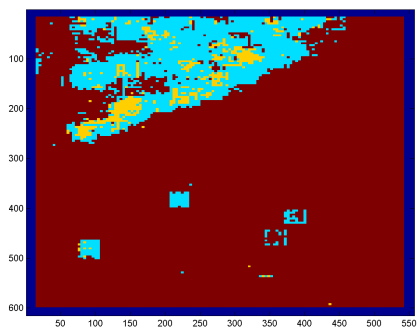}
\vspace{-0.5cm}
\caption{Automatic segmented image.}
\vspace{-0.5cm}
\label{autoSegIm}
\end{figure}

Figure \ref{autoSegIm} shows the result of the classification of the same image than the one given in the figure \ref{expert}. Sand (in red) and rock (in blue) are quite well classified but ripple (in yellow) is not well segmented. The dark blue corresponds to that part of the image that was not considered for the classification.

Just by looking this figure \ref{autoSegIm} we cannot say whether the classification is good or not, and any decision stays very subjective. Moreover, the classification algorithm could be good for this image and not for others. So we propose to use our measures. The used weights here for the certitude are respectively 2/3 for sure, 1/2 for moderately sure and 1/3 for not sure. But other weights can be preferred according to the application.

The normalized confusion matrix obtained for one randomly partition of the database is given by:
\begin{equation}
\label{confusionmatrix}
\left(
\begin{array}{ccc}
40.51 & 5.77 & 53.72 \\
19.65 & 18.79 & 61.56 \\
3.51 & 1.15 & 95.34 \\
45.96 & 12.47 & 41.57 \\
\end{array}
\right)
\end{equation}
The last line means that there is shadow or other parts classified in class 1, 2 or 3. We can note that a high proportion of the rock or cobble (class 1) is classified as sand or silt (class 3), and most of the ripple (class 2) also. Sand and silt, the most common kinds of sediments on our images, are very well classified. The vector of good classification rate given by [40.51 18.79 95.34 0] and the vector of error classification rate given by [41.26 43.84 28.47 50.00] summarize these results. Whereas we have good classification for sand and silt, we also a lot of errors because other sediments are classified as sand or silt.

These results are not significant enough in order to well evaluate the obtained segmentation. Our proposed measures, given respectively by the equations (\ref{WDCN}) and (\ref{FD}) expressed in percentage, are 65.17 for the good detection criteria and 61.35 for the false alarm criteria, if we consider the direction based on the GVF the proposed measures give 63.11 for good detection criteria and 64.84 for the false alarm criteria. 

To better illustrate these two last measures, we have proceeded to four more randomly partitions. We obtain a mean of 63.53 for the good detection criteria with 3.37 for the standard deviation and a mean of 60.53 for the false alarm criteria with 7.72 for the standard deviation. If we consider the direction based on the GVF, we obtain a mean of 60.09 for the good detection criteria with 3.13 for the standard deviation and a mean of 52.62 for the false alarm criteria with 8.04 for the standard deviation. The standard deviations show that the good detection criteria is more stable than the false alarm criteria. Our two measures can well evaluate the good detection and the false alarm. When we consider the direction based on the GVF, the criteria decrease because of the weights given by the directions. Here, the deducted segmentation is dependent of the size of the tile, in this case it could be better to not consider the direction based on the GVF.

In order to evaluate the classifier approach, all these measures have to be compared to the same measures calculated for other parameterizations or for other classifier algorithms.

\section{CONCLUSION}

We have proposed some new evaluation measures for image classification and segmentation in uncertain environments. These new evaluation measures can take into account the uncertain labels. The proposed classification evaluation can be used for every kind of uncertain elements classification and our segmentation evaluation can be used for all image segmentation approaches. We have shown that a global image classification evaluation must be made by the evaluation of the classification and, at the same time, by the evaluation of the produced segmentation. The proposed confusion matrix take into account the uncertainty of the expert and also the inhomogeneous units ({\it e.g.} tiles in the case of local image classification). Moreover we have defined good classification and error classification rates from our confusion matrix. The proposed segmentation evaluation considers good and false detection boundary measures where the subjectivity of the expert is considered by the given uncertainty on the boundaries.

The fusion of the information provided by various experts in our proposed evaluation approach is made after an individual evaluation, which means that we fuse our different measures calculated for each expert. This fusion is made by using a simple sum: the uncertainty is considered directly in our measures. We can imagine fusing the information provided by experts before the evaluation in order to obtain uncertain and/or imprecise reality ({\it e.g.} defining fuzzy zones around the boundaries according to the certainty given by the experts). The fusion can be made also by belief functions defined from the uncertainties. In this case we have to redefine our proposed measures. For instance, the reality obtained by the fusion of experts could be used to outperform the learning step of the classification.

\end{document}